\documentclass[runningheads]{llncs}

\usepackage{graphicx}
\usepackage{amsmath}
\usepackage{amssymb}
\usepackage{wrapfig}
\usepackage{float}
\usepackage[hidelinks]{hyperref}
\usepackage{booktabs}
\usepackage{tabularx}
\usepackage{amssymb}

\title{Tensor Channel Equivariant Graph Neural Networks for Molecular Polarizability Prediction}

\titlerunning{Tensor Channel Equivariant GNNs}
\author{
Jean Philip Filling \and
Daniel Franzen \and
Michael Wand
}

\authorrunning{Filling et al.}

\institute{
Institute for Computer Science, Johannes Gutenberg University Mainz, Germany\\
\email{jefillin@uni-mainz.de}
}

\begin{document}

\maketitle

\begin{abstract}
We introduce a tensor-channel equivariant graph neural network for direct prediction of molecular polarizability tensors. Building on the efficient PaiNN architecture, we augment the hidden representation with explicit symmetric rank-2 tensor channels aligned with the decomposition of polarizability into isotropic and anisotropic components. In contrast to approaches that construct tensor outputs only at readout, our model propagates tensor structure throughout message passing using geometrically motivated tensor bases. This yields a target-aligned architecture for tensor-valued molecular prediction.

On optimized QM7-X geometries, the proposed model achieves lower full-tensor and anisotropic error than both a PaiNN-style readout baseline and a dielectric MACE baseline under matched training conditions and at nearly identical parameter count. In this controlled setting, it also outperforms MACE while remaining substantially faster at inference. Ablation studies show that the gain does not arise from increased capacity alone, but from the combination of explicit tensor propagation and a traceless target parameterization matched to the anisotropic part of the polarizability tensor. Among the tensor bases considered, the strongest results are obtained from interactions between learned directional features, indicating that these are particularly effective for modeling molecular polarizability. Rotational equivariance tests further confirm that all compared models are numerically equivariant, so the observed improvements are attributable to better learning of the target tensor itself. Overall, our results show that for structured tensor-valued targets, propagating target-aligned tensor features can outperform both readout-only tensor construction and a more general higher-order equivariant model in the present training setting.
\end{abstract}

\section{Introduction}

Molecular polarizability describes how a molecule's electronic density responds to an external electric field and plays a central role in intermolecular interactions, dielectric behavior, and molecular materials modeling. Reliable polarizability tensors can be obtained from density functional theory (DFT), but the associated computational cost limits their use in large-scale simulations and high-throughput settings. This motivates the development of learned surrogate models for tensor-valued molecular properties.

For atomistic machine learning, respecting geometric symmetries is essential. Predictions should transform predictably under rotations, a property captured by equivariance. While equivariance can be encouraged through data augmentation, architectures that enforce it directly are typically more data-efficient \cite{cohen2016groupequivariantconvolutionalnetworks,cohen2019gaugeequivariantconvolutionalnetworks,franzen2021general}. Equivariant graph neural networks (GNNs) have therefore become a standard tool for molecular prediction tasks and have shown strong performance across a wide range of atomistic benchmarks \cite{schutt2017schnet,unke2019physnet,gasteiger2020fast}.

Most existing atomistic models are designed primarily around scalar targets. Tensor-valued observables, by contrast, are often handled indirectly or only at the level of the output head. For molecular polarizability, this raises a more specific question: is it sufficient to construct the target tensor only at readout, or does prediction benefit from propagating tensor structure explicitly throughout the network? We study this question in the setting of molecular polarizability prediction.

To this end, we start from PaiNN, a simple and efficient equivariant architecture, and extend it with learned symmetric rank-2 tensor channels. The design is aligned with the natural decomposition of the polarizability tensor into isotropic and anisotropic components, allowing the model to represent tensor structure directly in its hidden states rather than introducing it only at the final output stage. This yields a controlled test of whether explicit tensor propagation improves the learning of structured tensor-valued targets.

Our contributions are as follows:
\begin{itemize}
    \item We introduce a tensor-channel extension of PaiNN that propagates symmetric rank-2 tensor features throughout message passing.
    \item We use a target-aligned parameterization of molecular polarizability that separates isotropic and anisotropic structure and enforces tracelessness in the anisotropic component.
    \item We show in a controlled comparison on optimized QM7-X geometries that explicit tensor propagation improves full-tensor and especially anisotropic prediction over a readout-only PaiNN baseline, while also outperforming a dielectric MACE baseline at substantially lower inference cost.
\end{itemize}

These results indicate that, for structured tensor-valued molecular targets, the relevant question is not only whether the output is tensorial, but whether tensor structure is represented and propagated internally throughout the model.

\section{Related Work}
Equivariant neural networks for molecular systems are designed to combine invariant atomic features, such as element type, charge, or spin, with directional information extracted from molecular geometry. The central challenge is to couple these quantities so that the resulting computation respects rotational symmetry, meaning that messages and features transform consistently under rotations. A common solution is to replace a direct real-space formulation by a representation-theoretic one, where features are organized in irreducible representations of the rotation group. In this representation, rotations are handled conveniently through Wigner matrices, while interactions between features are realized by tensor products governed by Clebsch--Gordan coefficients. Early tensor field network approaches established this paradigm and showed how equivariant message passing can be formulated directly in terms of spherical harmonics and representation coupling \cite{thomas2018tensor,fuchs2020se,batzner20223,fu2025learning}. This perspective has since given rise to highly expressive architectures for atomistic modeling. A prominent example is MACE \cite{batatia2022mace,batatia2026mace}, which propagates features of multiple angular orders and achieves strong performance through repeated equivariant tensor product interactions. However, this expressivity typically comes at the cost of substantial architectural and computational complexity.

A related direction aims to retain the expressive power of tensor interactions while moving from irreducible representations to Cartesian tensor representations. TensorNet \cite{simeon2023tensornet} follows this idea and constructs an $\mathrm{O}(3)$-equivariant architecture directly in Cartesian coordinates, where feature transformations can be implemented through matrix and tensor operations rather than explicit Clebsch--Gordan coupling. This makes tensor interactions more accessible while preserving a principled equivariant structure. However, these approaches still rely on a fairly general tensor backbone and correspondingly rich tensor interactions.

This raises a natural question: can tensor-valued molecular properties be learned with a simpler architecture if more structure is built into the parameterization of the target quantity itself? One route in this direction is to use local coordinate frames, which allow tensorial information to be expressed in a more structured and task-oriented way. Recent work has shown that tensorial messages exchanged between local frames can provide an effective mechanism for communicating geometric information \cite{lippmann2024beyond,gerhartz2025equivariance}. At the same time, frame-based approaches introduce additional design choices, since the construction of local reference frames is not always unique and may become ambiguous in highly symmetric or nearly degenerate geometries.

\cite{filling2025direct} followed this local-frame perspective and introduced an equivariant architecture for predicting molecular polarizability tensors through tensor-valued message passing in local coordinate systems. That work demonstrated the promise of direct tensor regression for polarizability prediction, but it also highlighted the practical challenges associated with local frame construction.

In parallel, a different and highly successful design philosophy was established by PaiNN \cite{schutt2021equivariant}. Instead of relying on irreducible representations and explicit tensor products, PaiNN maintains equivariance using only scalar and vector channels, with directional information injected through interatomic displacement vectors. This yields a comparatively simple and efficient architecture that has proven effective across a range of atomistic tasks. However, tensor-valued quantities are typically not propagated as dedicated hidden representations throughout message passing.

Our work is most closely related to this scalar--vector line of models. Starting from the simplicity of PaiNN, we extend the architecture with explicit symmetric rank-2 tensor channels tailored to the prediction of tensorial quantities, like molecular polarizability tensor. In this sense, our approach can be viewed as a middle ground between highly expressive but more elaborate tensor-product architectures and simpler scalar--vector models. We retain the accessibility of the latter while introducing task-specific tensor representations directly into the message passing process.

\section{Molecular Polarizability}

Molecular polarizability describes the linear response of a molecule to a homogeneous external electric field $\mathbf{F}$. For sufficiently small fields, the induced dipole moment $\Delta\boldsymbol{\mu}=\boldsymbol{\mu}(\mathbf{F})-\boldsymbol{\mu}^{(0)}$ satisfies
\[
\Delta \mu_i = \sum_j \alpha_{ij} F_j,
\qquad
\alpha_{ij} = - \left.\frac{\partial^2 E}{\partial F_i \partial F_j}\right|_{\mathbf{F}=0}.
\]
Hence, $\boldsymbol{\alpha}=\boldsymbol{\alpha}(\mathbf{R})$ is a geometry-dependent second-order tensor, where $\mathbf{R}$ denotes the molecular geometry. Under a rotation $R\in\mathrm{SO}(3)$, it transforms as $\boldsymbol{\alpha}'=R\boldsymbol{\alpha}R^\top$, and more generally under $Q\in O(3)$ as $\boldsymbol{\alpha}'=Q\boldsymbol{\alpha}Q^\top$.

Since $\boldsymbol{\alpha}$ is symmetric, it contains six independent components and can be decomposed into an isotropic scalar part and a traceless anisotropic part,
\[
\alpha_{\mathrm{iso}}=\frac{1}{3}\mathrm{Tr}(\boldsymbol{\alpha}),
\qquad
\boldsymbol{\alpha}_{\mathrm{aniso}}=\boldsymbol{\alpha}-\alpha_{\mathrm{iso}}I,
\qquad
\boldsymbol{\alpha}=\alpha_{\mathrm{iso}}I+\boldsymbol{\alpha}_{\mathrm{aniso}}.
\]

\begin{figure}[H]
    \centering
    \includegraphics[width=0.52\textwidth]{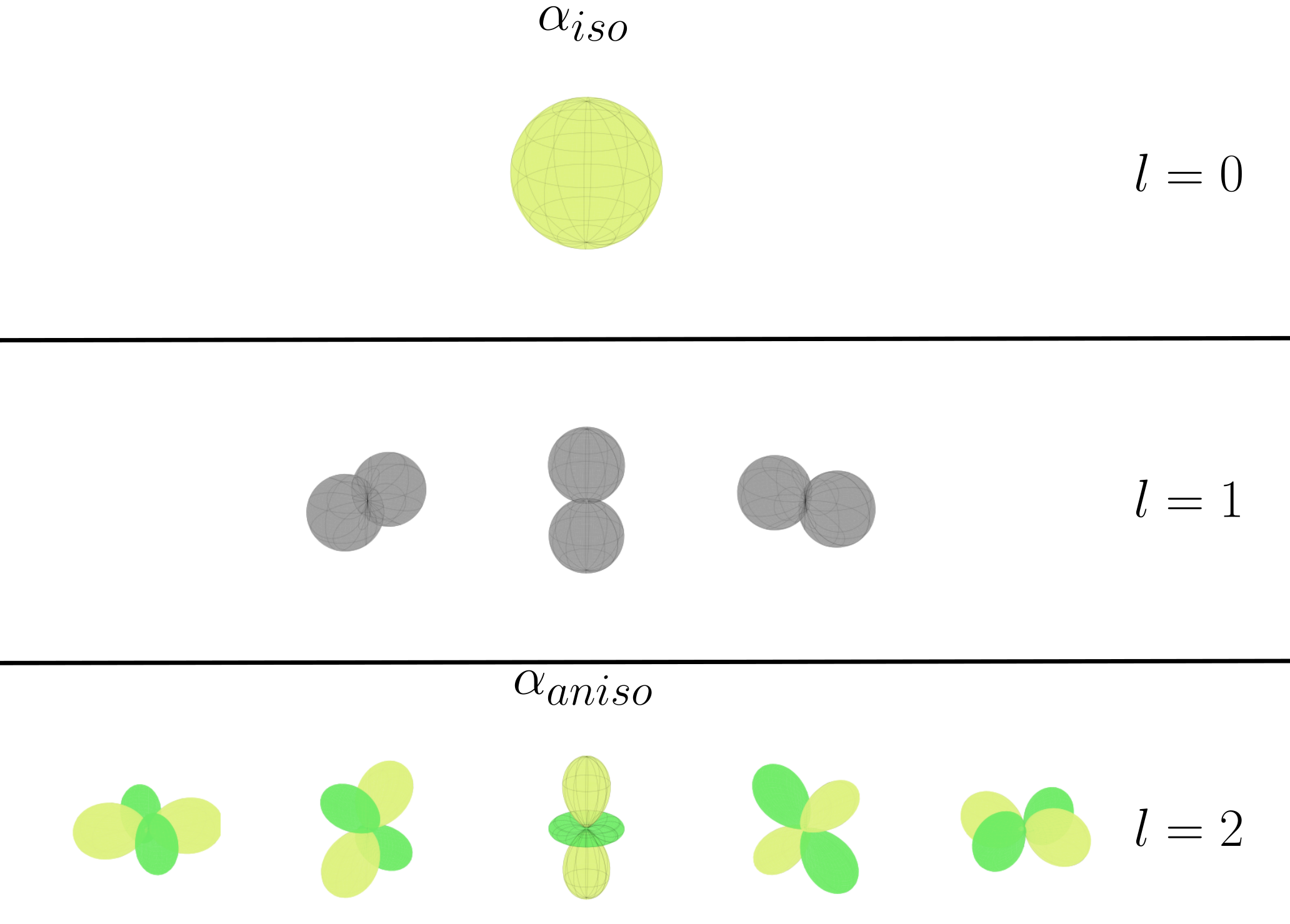}
    \caption{Spherical-harmonic decomposition of a symmetric second-order tensor. For the polarizability tensor, only the isotropic $\ell=0$ and the traceless symmetric anisotropic $\ell=2$ components remain.}
    \label{fig:sh}
\end{figure}

In the spherical-harmonic basis shown in Fig.~\ref{fig:sh}, $\alpha_{\mathrm{iso}}$ corresponds to the $\ell=0$ component, while $\boldsymbol{\alpha}_{\mathrm{aniso}}$ corresponds to the $\ell=2$ component. This is consistent with
\[
1 \otimes 1 = 0 \oplus 1 \oplus 2,
\]
where $\otimes$ denotes the dyadic (tensor) product and $\oplus$ the direct-sum decomposition into irreducible spherical-harmonic components. The $\ell=1$ contribution corresponds to the antisymmetric part and therefore vanishes for the symmetric polarizability tensor. In PaiNN, atomic features are represented by scalar ($\ell=0$) and vector ($\ell=1$) channels. Extending this representation by second-order tensor channels therefore provides a natural way to model molecular polarizability.

\section{Method}
Each atom $i$ carries scalar, vector, and tensor states,
$s_i \in \mathbb{R}^{C_s}$, $v_i \in \mathbb{R}^{C_v \times 3}$, and
$T_i \in \mathbb{R}^{C_t \times 3 \times 3}$.
For each edge $(j \to i)$, we define
$r_{ij}=x_j-x_i$, $d_{ij}=\|r_{ij}\|$, and $\hat r_{ij}=r_{ij}/d_{ij}$.
In addition, we expand pairwise distances using radial basis features
$e_{ij}^{\mathrm{rbf}} = \mathrm{RBF}(d_{ij})$, where $\mathrm{RBF}$ denotes a set of Bessel basis functions. $\odot$ a gate, that stands for a scalar-vector or scalar-tensor multiplication and $\otimes$ is the dyadic product.

\begin{figure}[!htbp]
  \centering
  \vspace{-0.5em} % verkleinert Abstand nach oben
  \includegraphics[width=\linewidth]{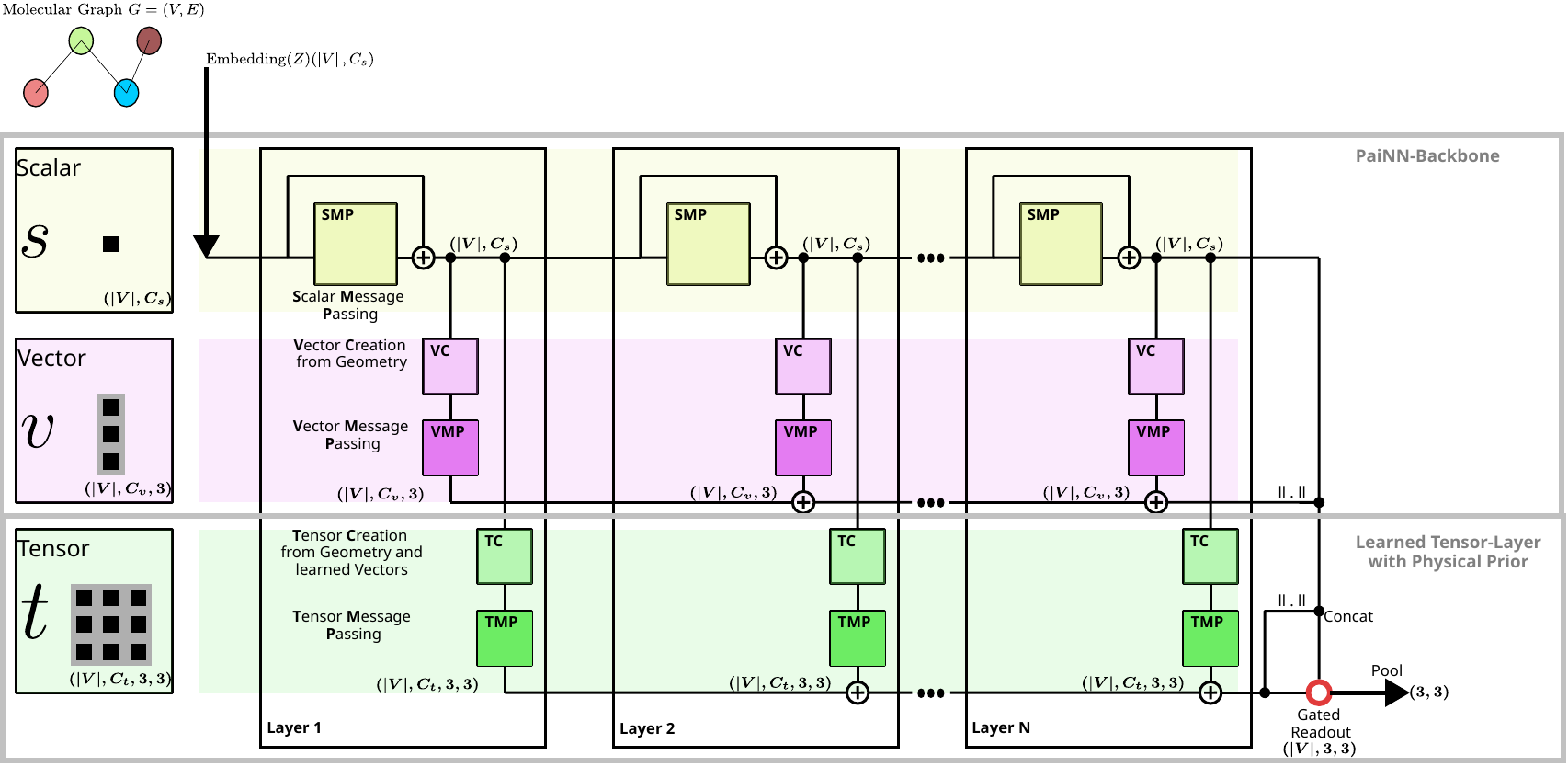}
  \vspace{-0.5em} % verkleinert Abstand vor caption
  \caption{\textbf{Equivariant Architecture for Rank-2 Tensors}}
  \vspace{-0.75em} % verkleinert Abstand nach caption
  \label{fig:architecture}
\end{figure}

\subsection{Scalar Channel}
The scalar channel performs invariant message passing in the spirit of EGNN-style \cite{satorras2021n} updates, where messages depend only on invariant quantities. In our case, the scalar message is defined as
\[
m_{ij}^{(s)} = \phi_s\!\left(
s_i \mathbin{\|} s_j \mathbin{\|} e_{ij}^{\mathrm{rbf}} \mathbin{\|} \psi(v_i, v_j, t_i, t_j)
\right),
\]
where \(\psi(v_i, v_j, t_i, t_j)\) denotes invariant summaries of the vector and tensor features. The aggregated message is summed over the local neighborhood and used in a residual update,
\[
s_i' = s_i + \phi_s^{\mathrm{upd}}\!\left(s_i \mathbin{\|} \sum_{j\in\mathcal N(i)} m_{ij}^{(s)}\right).
\]
This allows the scalar channel to collect local geometric information while remaining fully invariant.

\subsection{Vector Channel}
The vector channel follows the general design philosophy of PaiNN \cite{schutt2021equivariant}, combining directional information from interatomic displacements with learned transformations of previous vector features. The vector message is written as
\[
m_{ij}^{(v)} =
g_{ij}\Big(a_{ij}\hat r_{ij} + \phi_{\mathrm{mix}}(s_i \mathbin{\|} s_j \mathbin{\|} e_{ij}^{\mathrm{rbf}},\, v_j)\Big),
\]
where the first term injects directional information through \(\hat r_{ij}\) and the second term propagates learned vector features. The node update is given by
\[
v_i' = v_i + \phi_v^{\mathrm{gate}}(s_i)\odot \sum_{j\in\mathcal N(i)} m_{ij}^{(v)} .
\]
In this way, the vector channel retains the simplicity of PaiNN while providing a natural interface between invariant scalar information and directional geometric structure.

\subsection{Tensor Channel}
To propagate tensor-valued information explicitly, we introduce a tensor channel based on three branch-specific tensor bases,
\[
RR:\ \hat r_{ij}\otimes \hat r_{ij},\qquad
RV:\ \mathrm{sym}(\hat r_{ij}\otimes u_j),\qquad
VV:\ \mathrm{sym}(u_j\otimes w_j),
\]
where \(\mathrm{sym}(A)=\tfrac12(A+A^\top)\). In all three branches, we use the symmetric traceless part of the corresponding basis tensor,
\[
\mathrm{TL}(A)=\mathrm{sym}(A)-\frac{\mathrm{tr}(A)}{3}I,
\]
so that the tensor channel explicitly parameterizes the anisotropic contribution. The tensor update is
\[
t_i' = t_i + \phi_t^{\mathrm{gate}}(s_i)\odot
\sum_{j\in\mathcal N(i)}
\big(m_{ij}^{RR}+m_{ij}^{RV}+m_{ij}^{VV}\big).
\]
The \(RR/RV/VV\) decomposition is motivated by the different sources of anisotropy it captures: \(RR\) encodes purely geometric anisotropy, \(RV\) couples geometry with learned directional features, and \(VV\) models higher-order interactions among latent directional channels. Using a symmetric traceless basis is particularly natural for molecular polarizability, where the trace corresponds to the isotropic average response, while the traceless part captures the directional anisotropy.

Our model predicts an atom-wise polarizability tensor as a learned gated nonlinear superposition \cite{weiler20183d} of the propagated tensor channels and obtains the molecular tensor by additive pooling over all atoms. In contrast, the PaiNN-style baseline introduces tensor structure only in the final readout, where each atomic contribution is written as \(\alpha_i = \alpha_0(s_i) I + \mathrm{sym}(\nu_i \otimes r_i)\) with an isotropic scalar term \(\alpha_0(s_i) I\) and an anisotropic part derived from vector features. For each channel, skip connections are added for smoothing the loss landscape \cite{mehmeti2020ringing}.

\section{Experiment}

\textbf{Dataset}
We use the QM7-X dataset \cite{hoja2021qm7}, which contains about 4.2 million equilibrium and non-equilibrium structures of small organic molecules composed of H, C, N, O, S, and Cl. Reference properties are provided at the PBE0+MBD level, while the equilibrium geometries are obtained from DFTB3+MBD structure optimization. Because molecular polarizability is directly linked to dielectric and optical response, QM7-X provides a suitable basis for studying ML models for molecular materials design. In our preprocessing, we retain only optimized geometries ($\approx42,000$), remove duplicates, and store atomic numbers, Cartesian coordinates, and molecular polarizability tensors in a compact \texttt{meta.pickle} file. Train/validation/test splits are generated at the molecule level by grouping entries by \texttt{mol\_id}, shuffling molecule IDs with seed 42, and assigning them to 80\%, 10\%, and 10\% partitions.

\subsection{Metrics and Implementation Details}

\textbf{Metrics}
We evaluate predicted polarizability tensors using three complementary error measures.
Let $\hat{\alpha}_i$ and $\alpha_i$ denote the predicted and reference polarizability
tensors for molecule $i$. As the primary metric, we report the mean Frobenius error on
the full tensor, which measures the overall prediction quality. To separately assess the
isotropic contribution, we report the mean absolute error of the one-third trace,
$\frac{1}{3}\mathrm{Tr}(\alpha)$. To assess the anisotropic contribution, we report the
Frobenius error on the deviatoric part,
$\mathrm{dev}(\alpha)=\alpha-\frac{1}{3}\mathrm{Tr}(\alpha)I$.
Together, these metrics allow us to distinguish overall tensor accuracy from errors in
the isotropic and traceless anisotropic components. For each configuration, model
selection is performed using validation Frobenius MAE, and we report the corresponding
test metrics.

\subsection{Ablation Study}

We ablate the tensor channel to identify which design choices drive the performance gain. RR, RV, and VV represent purely geometric anisotropy, interactions between geometry and learned directional features, and interactions among learned directional features, respectively. We additionally study \textbf{SYM}, which enforces tensor symmetry, \textbf{TL}, which removes the trace and restricts the tensor channel to the anisotropic part, and \textbf{LoRA}, which performs channel mixing efficiently via a low-dimensional latent space. All variants are trained with the same optimization setup and evaluated using the Frobenius MAE on the full tensor, the MAE of the one-third trace, and the Frobenius MAE on the deviatoric part.

\begin{table}[h]
\centering
\caption{Ablation study within the tensor channel family on QM7-X. Lower is better for all metrics.}
\label{tab:ablation}
\begin{tabular}{l|ccc|ccc|c|ccc}
\toprule
Description & RR & RV & VV & Sym & TL & LoRa & Train Params & $\|\Delta \alpha\|_F$ & $|\Delta \alpha_{\mathrm{iso}}|$ & $\|\Delta \alpha_{\mathrm{aniso}}\|_F$ \\
\midrule
RR RV VV             & \checkmark & \checkmark & \checkmark & \checkmark & \checkmark & \checkmark & $8.71 \times 10^{6}$  & 0.157 & 0.043 & 0.129 \\
RR only              & \checkmark & --         & --         & \checkmark & \checkmark & --         & $5.43 \times 10^{6}$ & 0.212 & 0.048 & 0.186 \\
RV only              & --         & \checkmark & --         & \checkmark & \checkmark & \checkmark & $5.41 \times 10^{6}$ & 0.155 & 0.040 & 0.130 \\
RV trace feedback    & --         & \checkmark & --         & \checkmark & \checkmark & \checkmark & $5.64 \times 10^{6}$ & 0.154 & 0.040 & 0.128 \\
VV only              & --         & --         & \checkmark & \checkmark & \checkmark & \checkmark & $5.42 \times 10^{6}$ & \textbf{0.150} & \textbf{0.040} & \textbf{0.122} \\
VV w/o TL op.        & --         & --         & \checkmark & \checkmark & --         & \checkmark & $5.42 \times 10^{6}$ & 0.158 & 0.042 & 0.131 \\
VV w/o SYM TL op.    & --         & --         & \checkmark & --         & --         & \checkmark & $5.42 \times 10^{6}$ & 0.159 & 0.043 & 0.131 \\
VV w/o LoRa          & --         & --         & \checkmark & \checkmark & \checkmark & --         & $5.42 \times 10^{6}$ & 0.156 & 0.043 & 0.128 \\
\bottomrule
\end{tabular}
\end{table}

Table~\ref{tab:ablation} shows a clear pattern. RR performs substantially worse than RV and VV, indicating that geometry alone is not sufficient, while coupling tensor construction to learned directional features is crucial. The best results are obtained with VV, whereas combining all branches does not improve over VV alone despite the larger model. Removing TL or SYM degrades performance, confirming that the gain comes from a target-aligned tensor parameterization rather than from added tensor capacity alone. LoRA gives a small additional benefit while making channel mixing more efficient, whereas trace feedback has only a marginal effect. We therefore use the constrained VV variant with SYM, TL, and LoRA in the following experiments.

\section{Benchmark}

To test whether the gain of the tensor channel model comes from the architecture rather than from model size or optimization, we compare against two strong baselines at nearly identical parameter count and under matched training conditions: a PaiNN-style model that constructs the tensor only in the final readout, and a dielectric MACE model as a higher-order equivariant baseline. We report the Frobenius MAE of the full polarizability tensor, the MAE of the isotropic contribution given by one-third of the trace, and the Frobenius MAE of the anisotropic contribution.

\begin{table}[h]
\centering
\caption{Benchmark comparison on QM7-X. Lower is better for all error metrics and for runtime. Error metrics are reported as mean $\pm$ standard deviation over 5 independent training seeds (0, 21, 42, 63, 84) in Bohr$^3$, while runtime is reported in milliseconds per sample for one forward pass on a RTX4090.}
\label{tab:baseline_benchmark}
\begin{tabular}{lccccc}
\toprule
Model & Train Params & $\|\Delta \alpha\|_F$ & $|\Delta \alpha_{\mathrm{iso}}|$ & $\|\Delta \alpha_{\mathrm{aniso}}\|_F$ & ms/sample \\
\midrule
MACE        
& $5.41 \times 10^{6}$ 
& $0.197 \pm 0.007$ 
& $0.045 \pm 0.001$ 
& $0.170 \pm 0.007$ 
& 1.025 \\

PaiNN-style
& $5.42 \times 10^{6}$ 
& $0.164 \pm 0.002$ 
& $\mathbf{0.038 \pm 0.001}$ 
& $0.142 \pm 0.002$ 
& \textbf{0.392} \\

Ours (VV only)       
& $5.42 \times 10^{6}$ 
& $\mathbf{0.154 \pm 0.002}$ 
& $0.041 \pm 0.001$ 
& $\mathbf{0.127 \pm 0.002}$ 
& 0.511 \\
\bottomrule
\end{tabular}
\end{table}

Table~\ref{tab:baseline_benchmark} shows a clear ranking across seeds. Our model achieves the lowest error on the full tensor and on the anisotropic component, while the PaiNN-style baseline remains best on the isotropic term. Relative to PaiNN-style, the full-tensor error decreases from $0.164$ to $0.154$ and the anisotropic error from $0.142$ to $0.127$, whereas the isotropic error changes only slightly. The gain is therefore concentrated in the anisotropic contribution, indicating that tensor propagation mainly improves the modeling of directional structure rather than the trace component.

Compared with MACE, the gap is even larger: our model improves both full-tensor and anisotropic accuracy at essentially identical parameter count, while also showing smaller seed variance. In terms of runtime, PaiNN-style remains the fastest model, but our model adds only moderate cost relative to PaiNN-style and remains roughly twice as fast as MACE. Overall, these results show that explicitly propagating learned rank-2 tensor features provides a better inductive bias for molecular polarizability than readout-only tensor construction, and that this specialized architecture outperforms the more general MACE baseline under the training setting considered here.

\paragraph{Implementation Details.}
All models are trained on the same fixed molecule-level split with identical optimization settings, and model selection is based on validation Frobenius MAE. Unless noted otherwise, all models use a cutoff of 10\,\AA, 8 interaction layers, matched parameter counts, cosine learning-rate scheduling, the same EMA setup, and the same number of training epochs; all experiments are run on a single RTX4090. Our tensor channel model uses 148 scalar, 37 vector, and 37 tensor channels. The PaiNN-style baseline uses 156 scalar and 64 vector channels, reuses our scalar and vector interaction layers, and constructs the polarizability tensor only in the final PaiNN-style readout. For the dielectric MACE baseline, we use the authors official implementation and retrain it under the same setup. It uses 128 $\ell=0$, 128 $\ell=1$, and 128 $\ell=2$ channels \cite{batatia2022mace}.

\subsection{Rotational Equivariance Test}

We test rotational equivariance on the test set by applying 64 sampled proper rotations \(R\) to each molecule and reporting
\[
\varepsilon_{\mathrm{equiv}}=\left\| \hat\alpha(Rx)-R\hat\alpha(x)R^\top \right\|_F,
\qquad
\varepsilon_{\mathrm{target}}=\left\| \hat\alpha(Rx)-R\alpha(x)R^\top \right\|_F,
\]
where \(\varepsilon_{\mathrm{equiv}}\) measures rotational consistency and \(\varepsilon_{\mathrm{target}}\) the prediction error on rotated inputs. All values are averaged over test samples and rotations.

\begin{table}[h]
\centering
\caption{Rotational equivariance test on QM7-X using 64 sampled rotations. Lower is better for both metrics. All values are reported in Bohr$^3$. We used float32 models.}
\label{tab:rotation_benchmark}
\begin{tabular}{lcc}
\toprule
Model & $\varepsilon_{\mathrm{equiv}}$ & $\varepsilon_{\mathrm{target}}$ \\
\midrule
MACE baseline        & 0.000027 & 0.191077 \\
PaiNN-style baseline & 0.000025 & 0.166019 \\
Ours (VV only)       & 0.000028 & 0.150673 \\
\bottomrule
\end{tabular}
\end{table}

Table~\ref{tab:rotation_benchmark} shows that all models are numerically highly equivariant, with \(\varepsilon_{\mathrm{equiv}}\) on the order of \(10^{-5}\) Bohr$^3$. The differences in rotational consistency are therefore negligible. In contrast, \(\varepsilon_{\mathrm{target}}\) follows the same ranking as in the main benchmark, with our model achieving the lowest error. The performance differences are thus not caused by broken equivariance, but by differences in how accurately the polarizability tensor is learned.

\subsection{Error Analysis}

\begin{wrapfigure}{r}{0.48\textwidth}
    \centering
    \vspace{-0.5em}
    \includegraphics[width=0.46\textwidth]{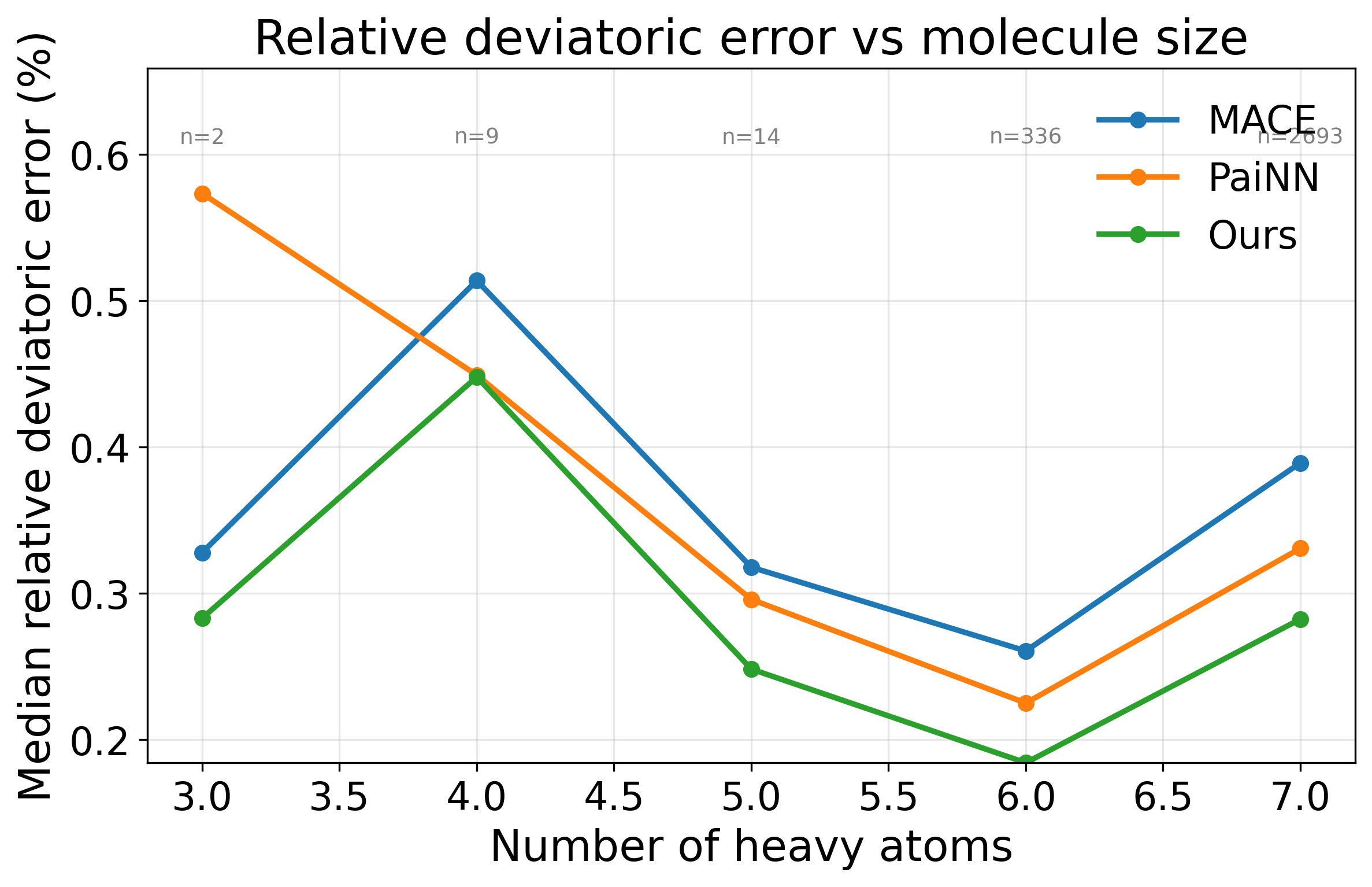}
    \caption{Median relative deviatoric error as a function of the number of heavy atoms. Numbers above the plot indicate the number of test samples in each bin.}
    \label{fig:rel_dev_err_heavy}
    \vspace{-1.0em}
\end{wrapfigure}

To further analyze the anisotropic prediction quality, we evaluate the relative deviatoric error as a function of molecular size. For a predicted polarizability tensor $\hat{\alpha}$ and target tensor $\alpha$, we define the relative deviatoric error as
\[
\frac{\|\mathrm{dev}(\hat{\alpha}) - \mathrm{dev}(\alpha)\|_F}{\|\mathrm{dev}(\alpha)\|_F + \varepsilon},
\]
where $\mathrm{dev}(\alpha)=\alpha-\tfrac{\mathrm{tr}(\alpha)}{3}I$ denotes the deviatoric part and $\varepsilon$ is a small constant for numerical stability. We group the test samples by the number of heavy atoms and report the median relative deviatoric error within each bin.

As shown in Figure~\ref{fig:rel_dev_err_heavy}, our model achieves the lowest relative deviatoric error across all heavy-atom bins. The same ordering is observed consistently for all sizes, with PaiNN performing better than MACE, but remaining above our model throughout. This indicates that the improvement is not restricted to a narrow size regime, but persists across the molecular sizes represented in QM7-X. The largest bins correspond to molecules with six and seven heavy atoms, where the same trend remains clearly visible. The smaller bins with three to five heavy atoms contain substantially fewer test samples and should therefore be interpreted with more caution, however, they do not contradict the overall pattern.

\section{Discussion and Conclusion}

Our results show that the main advantage of the proposed model is not additional
capacity, but a target-aligned tensor representation. The gain is concentrated
in the anisotropic part of the polarizability tensor, which indicates that
explicit tensor propagation helps primarily where directional tensor structure
matters most. Under matched training conditions and nearly identical parameter
count, this design outperforms both a PaiNN-style readout-only baseline and a
strong MACE baseline.

At the same time, the ablations make clear that the benefit is not obtained by
adding arbitrary tensor machinery. Performance depends on how the tensor channel
is parameterized, and the best results are achieved by the constrained \(VV\)
variant with traceless projection. This suggests that, for structured tensor
targets, the decisive factor is not maximal architectural generality, but the
right inductive bias.

We presented a simple equivariant GNN with explicit rank-2 tensor channels for
direct molecular polarizability prediction. The central finding is that
propagating learned tensor features is more effective than constructing the
output tensor only at the final readout. More broadly, the results suggest that
equivariant models for tensor-valued targets can be improved substantially by
encoding the structure of the target directly into the hidden representation.
Future work could include testing this principle in more complex application scenarios.

\section*{Acknowledgments}
This work was supported by the German Research Foundation (DFG) through the Collaborative Research Center/Transregio TRR 146 “Multiscale Methods for Complex Systems in Soft Matter”, Project B7.
\bibliographystyle{splncs04}
\bibliography{references}

@inproceedings{schutt2021equivariant,
  title={Equivariant message passing for the prediction of tensorial properties and molecular spectra},
  author={Sch{\"u}tt, Kristof and Unke, Oliver and Gastegger, Michael},
  booktitle={International conference on machine learning},
  pages={9377--9388},
  year={2021},
  organization={PMLR}
}

@article{batatia2022mace,
  title={MACE: Higher order equivariant message passing neural networks for fast and accurate force fields},
  author={Batatia, Ilyes and Kovacs, David P and Simm, Gregor and Ortner, Christoph and Cs{\'a}nyi, G{\'a}bor},
  journal={Advances in neural information processing systems},
  volume={35},
  pages={11423--11436},
  year={2022}
}

@article{filling2025direct,
  title={Direct Molecular Polarizability Prediction with SO (3) Equivariant Local Frame GNNs},
  author={Filling, Jean Philip and Post, Felix and Wand, Michael and Andrienko, Denis},
  journal={arXiv preprint arXiv:2511.07087},
  year={2025}
}

@misc{cohen2016groupequivariantconvolutionalnetworks,
      title={Group Equivariant Convolutional Networks}, 
      author={Taco S. Cohen and Max Welling},
      year={2016},
      eprint={1602.07576},
      archivePrefix={arXiv},
      primaryClass={cs.LG},
      url={https://arxiv.org/abs/1602.07576}, 
}

@misc{cohen2019gaugeequivariantconvolutionalnetworks,
      title={Gauge Equivariant Convolutional Networks and the Icosahedral CNN}, 
      author={Taco S. Cohen and Maurice Weiler and Berkay Kicanaoglu and Max Welling},
      year={2019},
      eprint={1902.04615},
      archivePrefix={arXiv},
      primaryClass={cs.LG},
      url={https://arxiv.org/abs/1902.04615}, 
}

@article{schutt2017schnet,
  title={Schnet: A continuous-filter convolutional neural network for modeling quantum interactions},
  author={Sch{\"u}tt, Kristof and Kindermans, Pieter-Jan and Sauceda Felix, Huziel Enoc and Chmiela, Stefan and Tkatchenko, Alexandre and M{\"u}ller, Klaus-Robert},
  journal={Advances in neural information processing systems},
  volume={30},
  year={2017}
}

@article{unke2019physnet,
  title={PhysNet: A neural network for predicting energies, forces, dipole moments, and partial charges},
  author={Unke, Oliver T and Meuwly, Markus},
  journal={Journal of chemical theory and computation},
  volume={15},
  number={6},
  pages={3678--3693},
  year={2019},
  publisher={ACS Publications}
}

@article{gasteiger2020fast,
  title={Fast and uncertainty-aware directional message passing for non-equilibrium molecules},
  author={Gasteiger, Johannes and Giri, Shankari and Margraf, Johannes T and G{\"u}nnemann, Stephan},
  journal={arXiv preprint arXiv:2011.14115},
  year={2020}
}

@article{simeon2023tensornet,
  title={Tensornet: Cartesian tensor representations for efficient learning of molecular potentials},
  author={Simeon, Guillem and De Fabritiis, Gianni},
  journal={Advances in Neural Information Processing Systems},
  volume={36},
  pages={37334--37353},
  year={2023}
}

@article{thomas2018tensor,
  title={Tensor field networks: Rotation-and translation-equivariant neural networks for 3d point clouds},
  author={Thomas, Nathaniel and Smidt, Tess and Kearnes, Steven and Yang, Lusann and Li, Li and Kohlhoff, Kai and Riley, Patrick},
  journal={arXiv preprint arXiv:1802.08219},
  year={2018}
}

@article{lippmann2024beyond,
  title={Beyond canonicalization: How tensorial messages improve equivariant message passing},
  author={Lippmann, Peter and Gerhartz, Gerrit and Remme, Roman and Hamprecht, Fred A},
  journal={arXiv preprint arXiv:2405.15389},
  year={2024}
}

@article{gerhartz2025equivariance,
  title={Equivariance by Local Canonicalization: A Matter of Representation},
  author={Gerhartz, Gerrit and Lippmann, Peter and Hamprecht, Fred A},
  journal={arXiv preprint arXiv:2509.26499},
  year={2025}
}

@inproceedings{satorras2021n,
  title={E (n) equivariant graph neural networks},
  author={Satorras, V{\i}ctor Garcia and Hoogeboom, Emiel and Welling, Max},
  booktitle={International conference on machine learning},
  pages={9323--9332},
  year={2021},
  organization={PMLR}
}

@article{weiler20183d,
  title={3d steerable cnns: Learning rotationally equivariant features in volumetric data},
  author={Weiler, Maurice and Geiger, Mario and Welling, Max and Boomsma, Wouter and Cohen, Taco S},
  journal={Advances in Neural information processing systems},
  volume={31},
  year={2018}
}

@article{hoja2021qm7,
  title={QM7-X, a comprehensive dataset of quantum-mechanical properties spanning the chemical space of small organic molecules},
  author={Hoja, Johannes and Medrano Sandonas, Leonardo and Ernst, Brian G and Vazquez-Mayagoitia, Alvaro and DiStasio Jr, Robert A and Tkatchenko, Alexandre},
  journal={Scientific data},
  volume={8},
  number={1},
  pages={43},
  year={2021},
  publisher={Nature Publishing Group UK London}
}

@article{batzner20223,
  title={E (3)-equivariant graph neural networks for data-efficient and accurate interatomic potentials},
  author={Batzner, Simon and Musaelian, Albert and Sun, Lixin and Geiger, Mario and Mailoa, Jonathan P and Kornbluth, Mordechai and Molinari, Nicola and Smidt, Tess E and Kozinsky, Boris},
  journal={Nature communications},
  volume={13},
  number={1},
  pages={2453},
  year={2022},
  publisher={Nature Publishing Group UK London}
}

@article{fuchs2020se,
  title={Se (3)-transformers: 3d roto-translation equivariant attention networks},
  author={Fuchs, Fabian and Worrall, Daniel and Fischer, Volker and Welling, Max},
  journal={Advances in neural information processing systems},
  volume={33},
  pages={1970--1981},
  year={2020}
}

@article{franzen2021general,
  title={General nonlinearities in so (2)-equivariant cnns},
  author={Franzen, Daniel and Wand, Michael},
  journal={Advances in neural information processing systems},
  volume={34},
  pages={9086--9098},
  year={2021}
}

@article{fu2025learning,
  title={Learning smooth and expressive interatomic potentials for physical property prediction},
  author={Fu, Xiang and Wood, Brandon M and Barroso-Luque, Luis and Levine, Daniel S and Gao, Meng and Dzamba, Misko and Zitnick, C Lawrence},
  journal={arXiv preprint arXiv:2502.12147},
  year={2025}
}

@inproceedings{mehmeti2020ringing,
  title={Ringing relus: Harmonic distortion analysis of nonlinear feedforward networks},
  author={Mehmeti-G{\"o}pel, Christian HX Ali and Hartmann, David and Wand, Michael},
  booktitle={International Conference on Learning Representations},
  year={2020}
}

@article{batatia2026mace,
  title={MACE-POLAR-1: A Polarisable Electrostatic Foundation Model for Molecular Chemistry},
  author={Batatia, Ilyes and Baldwin, William J and Kuryla, Domantas and Hart, Joseph and Kasoar, Elliott and Elena, Alin M and Moore, Harry and Gawkowski, Miko{\l}aj J and Shi, Benjamin X and Kapil, Venkat and others},
  journal={arXiv preprint arXiv:2602.19411},
  year={2026}
}

\end{document}